\documentclass[11pt]{article}

\usepackage[final]{acl}
\usepackage{float}
\usepackage{times}
\usepackage{latexsym}
\usepackage{multirow}
\usepackage[T1]{fontenc}

\usepackage[utf8]{inputenc}

\usepackage{microtype}

\usepackage{inconsolata}
\usepackage{booktabs}
\usepackage{amsmath}
\usepackage{graphicx}
\usepackage{array}
\usepackage{listings}
\usepackage{fvextra}
\usepackage{color}
\usepackage{multirow}
\title{SarcasTraps: Effect of Sarcasm on Turkish Hate Speech Detection}

\author{
Niyazi Ahmet Metin \quad \quad Dilara Keküllüoğlu  \\
Sabanci University\\
Istanbul, Türkiye\\
\texttt{\{ahmet.metin,dilara.kekulluoglu\}@sabanciuniv.edu}
}

\begin{document}
\maketitle
\begin{abstract}

Standard hate speech detection datasets often emphasize explicit harmful language, but directed harm can also be expressed indirectly, without clearly hateful wording. Sarcasm is one such form of indirect expression, as it can convey harmful intent through mock praise, exaggeration, humor, or apparent politeness. In this paper, we introduce SarcasTraps, a targeted external test set for analyzing  the effect of sarcasm on Turkish hate speech detection. SarcasTraps contains 479 entries annotated along two dimensions: hate speech and sarcasm. We train classifiers using BERTurk and XLM-RoBERTa on two existing Turkish hate speech datasets and evaluate them on SarcasTraps, using it strictly as an external test set. We also evaluate GPT-4.1 and Llama-3.3-70B in a zero-shot setting on SarcasTraps. 
Results show that sarcasm substantially affects hate speech detection. Fine-tuned encoder models lose recall on sarcastic hate, while zero-shot LLMs maintain higher recall but suffer substantial precision drops on sarcastic examples. These findings show that both fine-tuned hate speech detectors and zero-shot LLMs struggle when sarcasm is involved, and SarcasTraps provides quantitative evidence of this challenge through a targeted Turkish test set. Due to the presence of sensitive and hateful language, SarcasTraps will be shared upon request for academic research purposes. \textcolor{red}{Warning: This paper has examples from the dataset which contains offensive and hateful content towards groups of people.}

\end{abstract}

\section{Introduction}

Understanding sarcasm is difficult, as even its definition is contested. Work in pragmatics repeatedly notes that the boundary between sarcasm and irony is elusive, and broader theories of irony have not settled on a single stable definition.~\citep{DYNEL201769,Kapogianni_2021}. Some sources take sarcasm as a subclass of irony~\citep{leggitt2000emotional}, and some count them as separate notions~\citep{ling2016empirical}. In general, sarcasm is described in terms of its overlap with irony rather than in isolation, in both, meanings inferred beyond the literal utterance, and there is a need to recover what a speaker intends rather than what the sentence literally says. Prior work on Turkish also suggests that the term ``sarcasm'' has no exact equivalent in Turkish; instead, several near-equivalents are used in everyday language. In Turkish literary tradition, sarcasm is most closely associated with ``hiciv''~\citep{10.1371/journal.pone.0276073}. Within pragmatics, sarcasm could be treated not as a simple gap between literal and intended meaning. Sarcasm is also a communicative resource since figurative language is used to accomplish discourse goals, and sarcastic irony is often used to complain or criticize implicitly while making the speaker appear less rude or unfair~\citep{richard_1994,JORGENSEN1996613}.

In Natural Language Processing, sarcasm has most commonly been studied as a text classification problem, where the goal is to determine whether a given text is sarcastic. Early work focused largely on English social media and microblog texts, and later extended to other genres such as sarcastic headlines and crowdsourced corpora~\citep{barbieri-etal-2014-modelling,ptacek-etal-2014-sarcasm,npapers_headlines,oraby-etal-2016-creating}. More recent work has also considered intended sarcasm, where labels are provided by the authors of the sarcastic texts themselves~\citep{isarcasm}. Beyond English, sarcasm and irony detection have been explored in languages such as Chinese and Arabic, as well as in multilingual settings including Arabic--English and Czech--English~\citep{chinese_2020,abu-farha-magdy-2020-arabic,abu-farha-etal-2022-sarcasm,ptacek-etal-2014-sarcasm}. In Turkish, existing resources have appeared under partially overlapping labels such as irony, satire, and sarcasm: earlier datasets focused on Twitter or microblog-style texts~\citep{taslioglu_karagoz,dulger,turkishmicroblogs_dl}, while later work introduced \textit{IronyTR}, Turkish satire datasets, and news-column datasets annotated for irony and sarcasm~\citep{cemek_ironytr,satire_ensemble,bilgin_metin}. Closest to our setting, \citet{metin-etal-2026-sarcasturk} introduced a context-aware Turkish sarcasm detection dataset from Ekşi Sözlük, using title-level contextual summaries to evaluate sarcasm detection in informal online discourse.

These studies provide the foundation for modeling sarcasm directly, but they also motivate asking how sarcasm interacts with other NLP tasks that depend on speaker intent. Thus, this study shifts the focus from sarcasm classification to investigating its broader effects; specifically, we examine whether sarcasm creates a hidden failure mode in the safety-critical domain of hate speech detection.

Hate speech detection is central to online safety, content moderation, and harmful language understanding~\citep{schmidt-wiegand-2017-survey}, yet many evaluation settings primarily evaluate cases where harmful intent is expressed relatively directly, through explicit insults, slurs, or negative statements about a target group.~\citep{waseem-hovy-2016-hateful,Davidson_Warmsley_Macy_Weber_2017,zampieri-etal-2019-predicting,elsherief-etal-2021-latent}. Sarcasm challenges this assumption because it can separate the literal wording of a text from its intended social meaning. A sarcastic utterance may appear humorous, polite, exaggerated, or indirectly critical on the surface while still conveying contempt, exclusion, or hostility toward a protected group. This motivates our central question: Do hate speech detectors trained on standard Turkish datasets remain reliable when hateful intent is expressed sarcastically? This problem is especially relevant in Turkish online discourse because informal platforms frequently contain indirect, mocking, and culturally loaded expressions.

To address this question, we introduce SarcasTraps where we analyze the effect of sarcasm on Turkish hate speech detection. Rather than treating sarcasm as the target label, SarcasTraps annotates each instance along two dimensions: whether the text contains hate speech and whether it is sarcastic. Combining these annotations yields four conditions: non-hateful non-sarcastic, non-hateful sarcastic, hateful non-sarcastic, and hateful sarcastic. In total, SarcasTraps contains 479 samples. We use SarcasTraps as an external test set by training hate speech classifiers on existing Turkish hate speech datasets and evaluating how their predictions transfer to these four classes.

We build SarcasTraps from Ekşi Sözlük (literally, ``Sour Dictionary''), one of the largest user-generated discussion platforms in Türkiye~\citep{eksisozluk}. The platform is organized around \textit{titles}, where each title names a concept, event, person, phrase, or social issue, and users contribute short \textit{entries} under that title. This structure produces a distinctive form of informal online discourse: entries are often subjective, culturally loaded, politically charged, humorous, or mocking, and their interpretation may depend on the shared topic established by the title. These properties make Ekşi Sözlük a suitable source for SarcasTraps, since it naturally contains cases where harmful group-level meaning can be expressed indirectly through sarcasm, irony, exaggeration, or mock praise. 
 
\section{Related Work}

Hate speech detection has been widely studied as a supervised classification problem, especially over social media texts. Early work introduced Twitter-based datasets and feature-based models for identifying hateful or offensive content, using lexical and surface-level signals such as character n-grams, indicative words, and hate speech keywords~\citep{waseem-hovy-2016-hateful,Davidson_Warmsley_Macy_Weber_2017}. Subsequent work expanded the task beyond binary hate detection by introducing large-scale abusive language datasets and more fine-grained annotation schemes for offensive content, including labels for offense type and target~\citep{founta-etal-2018-large,zampieri-etal-2019-predicting}. Recent work has also explored domain-adapted transformer encoders such as HateBERT for abusive and hate speech detection~\citep{caselli-etal-2021-hatebert}. However, recent work has noted that much hate speech detection research focuses on explicit or overt hate, leaving coded, indirect, or implicit forms less directly addressed~\citep{elsherief-etal-2021-latent}. This concern has motivated recent evaluation settings for implicit and adversarial hate speech, where harmful meaning can be expressed without overt hateful wording~\citep{elsherief-etal-2021-latent,hartvigsen-etal-2022-toxigen}. This line of work motivates our setting: if hate speech can be conveyed indirectly, then models trained on standard hate speech datasets may fail when group-directed hostility is expressed through non-literal forms such as sarcasm.

For Turkish, hate speech detection has been studied mostly through social media datasets and evaluation settings. Early Turkish resources include tweet-based corpora for offensive or abusive language detection, establishing initial baselines for harmful language identification in Turkish~\citep{coltekin-2020-corpus}. \citet{imayda} introduce a publicly available Turkish hate speech dataset of tweets annotated as hate, offensive, or none, with additional subclass labels for hateful examples such as ethnic, religious, sexist, and political hate. More recent work scales Turkish hate speech detection to larger and more domain-oriented settings: \citet{toraman-etal-2022-large} construct large-scale Turkish and English hate speech datasets and study cross-domain transfer, while \citet{suhs} provides Turkish hate speech benchmarks across target domains such as refugees, Greeks, and Israel--Palestine-related discourse. Beyond post-level classification, \citet{seker-etal-2025-hatecat} introduce HATECAT-TR, a span-annotated Turkish hate speech dataset that identifies hateful spans and categorizes them into fine-grained discourse types. These resources have substantially advanced Turkish hate speech detection, but they primarily model hate speech as post-level or span-level harmful language classification. In contrast, our work uses Turkish hate speech supervision to test whether such models remain reliable when hateful intent is expressed through sarcasm.

Recent work has increasingly examined how harmful language detection is affected when abuse or hate is expressed indirectly. \citet{frenda-etal-2023-when-sarcasm-hurts} investigates irony-aware models for abusive language detection, showing that incorporating irony and sarcasm information can improve the detection of abusive content, especially when harmful meaning is expressed indirectly. \citet{Frenda_Patti_Rosso_2023} shows that harmful content may appear with a weakly offensive surface form. Their analysis supports the idea that abusive intent can rely on indirect cues like sarcasm or irony. Some corpus construction efforts have also moved beyond single-label hate speech annotation by including additional pragmatic and discourse-level dimensions, including sarcasm. For example, \citet{fridriksdottir-etal-2024-ice} introduce Ice and Fire, an Icelandic blog-comment dataset annotated for multiple dimensions. Also, recent multi-dimensional social media datasets similarly annotate hate speech or offensive content together with related dimensions such as sarcasm, humor, stance, emotion, factuality, or spam~\citep{Gautam_Mathur_Gosangi_Mahata_Sawhney_Shah_2020,zaghouani-etal-2024-hateful,pramanik2026multisoc4dbenchmarkdiagnosinginstructioninduced}. 

\section{SarcasTraps}
\label{sec:sarcastraps}

To analyze the effect of sarcasm on Turkish hate detection, we curated the SarcasTraps dataset. All examples in SarcasTraps were collected from publicly accessible Ekşi Sözlük pages between April and June 2026. 

Candidate titles were identified using Turkish search terms related to Greek, Jewish/Israeli, and immigrant/refugee discourse. We purposively inspected titles whose framing made group-directed hostility reasonably likely, rather than inspecting all keyword matches. For example, a topic-neutral title such as \texttt{Syrian cuisine} was not selected, whereas a title framed around an \texttt{influx of Syrians} could be inspected.

The target domains were selected as they frequently appear in Turkish online discussions and have been used in Turkish hate speech studies and benchmarks~\citep{suhs,seker-etal-2025-hatecat,imayda}. The scope of SarcasTraps is limited to these domains. Entries whose hateful content targets only specific individuals are not included in the set. To protect privacy, individual person names occurring in the entries are masked using <PERSON\_n>.
In some discussions, users expressed hate and/or sarcasm in a specific sentence or phrase within a longer entry. In these cases, we extracted only the relevant portion containing the expression, rather than collecting the entire entry.

Three native Turkish annotators independently labeled each entry for sarcasm and hate speech. Two of the annotators are authors of this paper, while the third is an external annotator. The Krippendorff's $\alpha$ over the independently assigned labels was 0.755 for hate speech and 0.659 for sarcasm. After this independent annotation stage, the annotators met to discuss disagreements. Consensus was reached for all but five entries, whose final labels were determined through majority voting. During adjudication, four entries were removed because they did not satisfy the scope of the test set, including entries that expressed hostility toward a specific individual rather than a group-level identity. The final test set contains 479 entries. The final distribution consists of 124 non-hateful non-sarcastic, 107 non-hateful sarcastic, 154 hateful non-sarcastic, and 94 hateful sarcastic entries. 

Annotators were instructed to label an entry as sarcastic when its intended meaning differs from, reverses, or distorts its literal surface meaning. The sarcasm label includes mock politeness, mock approval, deliberate distortion, rhetorical questions, irony, parody, feigned ignorance, or absurd exaggeration. Prior work, ~\citep{oraby-etal-2016-creating}, has shown that these forms frequently participate in sarcastic or ironic utterances, while also occurring in non-sarcastic contexts.
These categories are not treated as separate annotation categories. The presence of any of these forms was treated as practical cues only when they supported a non-literal interpretation of the intended meaning.

Annotators were instructed to label an entry as hate speech if its intended meaning expresses a negative generalization, demeaning characterization, exclusionary stance, or hostile attitude toward a target group. The hate label was assigned only when the entry targeted a group-level identity or social category, rather than merely criticizing a specific individual.
For sarcastic examples, annotators were instructed to reconstruct the intended proposition behind the sarcastic surface form before assigning the hate label. 
For example, if an entry appears to praise a target group in a sarcastic way and the intended interpretation is the opposite of that praise (e.g., a demeaning or hostile claim about the group), the example is labeled as hate. 
In contrast, sarcastic examples are labeled as non-hate only when the reconstructed intended meaning does not contain group-directed hostility, degradation, exclusion, or harmful generalization.

Examples from all four combinations of the hate speech and sarcasm labels are presented in Table~\ref{tab:data_examples}. We report only English translations and partially mask profane expressions in the illustrative examples to reduce direct traceability and unnecessary exposure to harmful language. The underlying test set preserves the original wording, except that names of identifiable individuals are replaced with placeholders of the form \texttt{<PERSON\_n>}.

\begin{table*}[t!]
\centering
\setlength{\tabcolsep}{6pt}
\renewcommand{\arraystretch}{1}
\begin{tabular}{
  >{\raggedright\arraybackslash}m{0.26\linewidth}
  >{\raggedright\arraybackslash}m{0.62\linewidth}
  >{\centering\arraybackslash}m{0.04\linewidth}
  >{\centering\arraybackslash}m{0.04\linewidth}
}
\toprule
\textbf{Title} & \textbf{Entry} & \textbf{H} & \textbf{S} \\
\midrule

Turkish and Greek soldiers swearing at each other &
Made me laugh. Trash talk like this is better than actually fighting. &
0 & 0 \\
\midrule

Do the Freemasons or the Jews rule the world? &
Minibus drivers and taxi drivers rule the world. &
0 & 1 \\
\midrule

A message for Afghans and Pakistanis &
First of all, you are not our brothers. We don't have predatory,
dishonorable brothers like you. Leave. Get the f*** out of our country. &
1 & 0 \\
\midrule

The invasion of Caddebostan Beach by Syrians &
Oh, but they are Ansar; they are our brothers. What's the big deal if
they went into the sea in a chador, or if their men disturbed the peace
of society? There is a war! They got tired while fleeing the war;
shouldn't they enjoy the beach a little bit? &
1 & 1 \\

\bottomrule
\end{tabular}
\caption{Examples from the four conditions in
SarcasTraps. H indicates hate speech and S indicates sarcasm.}
\label{tab:data_examples}
\end{table*}

The non-sarcastic examples provide a reference condition for separating general domain-transfer effects from the additional effect of sarcasm.

\section{Experiments}
The purpose of our experiments is to test whether hate speech detectors trained on Turkish hate speech datasets remain reliable when hateful intent is expressed through sarcasm. To this end, we trained encoder-only classifiers using two Turkish hate speech datasets to which we obtained access. We evaluated the trained models both on the corresponding in-domain test sets and on the sarcastic and non-sarcastic subsets of SarcasTraps. We additionally evaluated two zero-shot large language models on SarcasTraps. The purpose of including multiple model configurations is not to establish a ranking across architectures, model scales, or training regimes. Instead, the models serve as complementary probes for examining whether changes across evaluation conditions are consistent across different system types. Accordingly, our analysis focuses on within-model performance differences rather than direct comparisons between fine-tuned encoders and zero-shot LLMs.

In the experiments, the models are evaluated as binary hate speech classifiers. That is, the target label is whether the input contains hate speech, while the sarcasm annotation is used only for analysis. SarcasTraps is used only as an external evaluation set, not for training or fine-tuning. 
 
To train hate speech detectors for SarcasTraps test evaluation, we used two existing Turkish hate speech datasets as training resources: the Mayda et al. dataset~\citep{imayda} and SU-HS~\citep{suhs}. 
The Mayda et al. dataset was obtained from the authors' public GitHub repository, where no explicit license was identified. The SU-HS dataset was obtained from the authors under a data-use agreement. Both datasets were used for academic research experiments, we cite the original creators and do not redistribute the datasets.
We selected these datasets because they both cover the target domains represented in SarcasTraps, including refugees/Syrians, Greeks, and Jewish/Israel--Palestine-related hate speech. This allows us to test whether models trained with standard hate speech supervision transfer to sarcastically expressed hate in the same broad target domains.
For Mayda et al., the original dataset contains three labels: hate, offensive, and none, and covers 22 target groups, including Syrian refugees, Greeks, and Jews. Since our experiments focus on binary hate speech detection, we excluded the offensive class and retained only the hate and none examples, resulting in 2,336 hate and 7,722 non-hate instances. We used examples from all 22 target groups in our experiments. As the dataset does not provide an official train/test split, we constructed a split where for each target group, we randomly sampled 10\% of the examples as the test set and used the remaining 90\% for training. This resulted in 9,051 training instances and 1,007 test instances.
For SU-HS, we used the dataset's provided binary train/test split without additional label filtering. The dataset covers four domains, including refugee-related hate speech, the Israel--Palestine conflict, and Greek-related hate speech. We used all available examples in these splits, resulting in 10,932 training instances and 2,743 test instances.

We trained separate encoder-based classifiers for each training dataset using BERTurk~\footnote{\url{https://huggingface.co/dbmdz/bert-base-turkish-128k-cased}}~\citep{schweter2020berturk} and XLM-RoBERTa~\footnote{\url{https://huggingface.co/xlm-roberta-base}}~\citep{conneau-etal-2020-unsupervised}. During evaluation on SarcasTraps, the model input is formed as \texttt{title + `` : '' + entry}. We did not perform hyperparameter search and fine-tuned all encoder models for five epochs with a batch size of 16, maximum sequence length of 384, weight decay of 0.01, warmup ratio of 0.06. We used a learning rate of $4 \times 10^{-5}$ for BERTurk and $2 \times 10^{-5}$ for XLM-RoBERTa. Each encoder configuration was trained with five random seeds: 42, 123, 456, 789, and 1024. Encoder fine-tuning was implemented with PyTorch~\citep{NEURIPS2019_bdbca288} and the Hugging Face Transformers Library~\citep{wolf-etal-2020-transformers}. Evaluation metrics were computed with scikit-learn~\citep{scikit-learn}. Fine-tuning was performed in Google Colab sessions, each using a single NVIDIA A100 GPU. Each SU-HS runs each consisted of 3,420 optimization steps, while each Mayda et al. runs each consisted of 2,830 steps. Across the four encoder configurations and five random seeds, this resulted in 20 fine-tuning runs. The total training time was approximately 1.8 GPU hours. After training, each classifier was first evaluated on the corresponding in-domain test set and then directly evaluated on SarcasTraps without any additional adaptation.

For decoder-only large language models, we performed zero-shot hate speech classification without fine-tuning. GPT-4.1~\citep{openai2025gpt41} and Llama-3.3-70B-Versatile~\citep{meta2024llama33} were evaluated using the same SarcasTraps instances. GPT-4.1 was accessed through the OpenAI API, and Llama-3.3-70B-Versatile was accessed through the Groq API.\footnote{\url{https://console.groq.com/docs/api-reference}} The models are prompted in Turkish and instructed to decide whether the given text contains hate speech. Hate speech is defined in the prompt as content that targets, demeans, threatens, or encourages violence against a person or group on the basis of identity-related attributes such as race, ethnicity, religion, gender, sexual orientation, disability, or similar characteristics. The prompt constrains the output to a JSON object with a binary hate label, together with a short justification. We use deterministic decoding with temperature 0. The full prompts are provided in Appendix~\ref{app:prompts}.

For all models, we report aggregate hate speech detection performance on the in domain test sets and hate-class precision, recall, and F1 on the non-sarcastic and sarcastic subsets of SarcasTraps. Since the purpose of SarcasTraps is to test whether hate speech detection systems behave differently when hate is expressed sarcastically, the main results are analyzed by comparing model behavior across non-sarcastic and sarcastic cases. Detailed performance patterns and error analyses are presented in Section~\ref{sec:results}.

\section{Results \& Analysis} \label{sec:results}
In this section, we evaluate model behavior across two successive conditions. We first examine whether the fine-tuned encoder models transfer from their corresponding in-domain test sets to the non-sarcastic subset of SarcasTraps. We then use this non-sarcastic subset as a reference condition and compare it with the sarcastic subset to analyze the effect of sarcasm on hate speech detection. 

Table~\ref{tab_0} compares the encoder models' performance on their corresponding in-domain test sets and on the non-sarcastic subset of SarcasTraps, denoted SarcasTraps (0). Zero-shot LLMs are not included in this comparison because they were not trained on either of the in-domain datasets and therefore do not have a corresponding in-domain evaluation condition.

\begin{table*}[t!]
\centering
\scriptsize
\setlength{\tabcolsep}{3.2pt}
\renewcommand{\arraystretch}{1.15}
\resizebox{\textwidth}{!}{
\begin{tabular}{l c c c c c c c c}
\toprule
\multirow{2}{*}{\textbf{Model}}
& \multicolumn{4}{c}{\textbf{In-domain Test}}
& \multicolumn{4}{c}{\textbf{SarcasTraps (0), $n=278$}} \\
\cmidrule(lr){2-5}
\cmidrule(lr){6-9}
& \textbf{Acc.} & \textbf{Prec.} & \textbf{Rec.} & \textbf{F1}
& \textbf{Acc.} & \textbf{Prec.} & \textbf{Rec.} & \textbf{F1} \\
\midrule

BERTurk$_{\text{SU-HS}}$
& 0.831{\scriptsize $\pm$ 0.010}
& 0.704{\scriptsize $\pm$ 0.017}
& 0.684{\scriptsize $\pm$ 0.058}
& 0.693{\scriptsize $\pm$ 0.030}
& 0.817{\scriptsize $\pm$ 0.009}
& 0.871{\scriptsize $\pm$ 0.016}
& 0.787{\scriptsize $\pm$ 0.011}
& 0.827{\scriptsize $\pm$ 0.007} \\

BERTurk$_{\text{Mayda}}$
& 0.940{\scriptsize $\pm$ 0.004}
& 0.886{\scriptsize $\pm$ 0.010}
& 0.852{\scriptsize $\pm$ 0.017}
& 0.869{\scriptsize $\pm$ 0.009}
& 0.812{\scriptsize $\pm$ 0.020}
& 0.838{\scriptsize $\pm$ 0.018}
& 0.818{\scriptsize $\pm$ 0.042}
& 0.828{\scriptsize $\pm$ 0.022} \\

XLM-RoBERTa$_{\text{SU-HS}}$
& 0.823{\scriptsize $\pm$ 0.005}
& 0.689{\scriptsize $\pm$ 0.016}
& 0.673{\scriptsize $\pm$ 0.013}
& 0.680{\scriptsize $\pm$ 0.007}
& 0.778{\scriptsize $\pm$ 0.028}
& 0.898{\scriptsize $\pm$ 0.020}
& 0.675{\scriptsize $\pm$ 0.046}
& 0.770{\scriptsize $\pm$ 0.034} \\

XLM-RoBERTa$_{\text{Mayda}}$
& 0.916{\scriptsize $\pm$ 0.006}
& 0.854{\scriptsize $\pm$ 0.009}
& 0.771{\scriptsize $\pm$ 0.040}
& 0.810{\scriptsize $\pm$ 0.019}
& 0.746{\scriptsize $\pm$ 0.027}
& 0.854{\scriptsize $\pm$ 0.034}
& 0.655{\scriptsize $\pm$ 0.041}
& 0.740{\scriptsize $\pm$ 0.031} \\

\bottomrule
\end{tabular}
}
\caption{Encoder performance on the corresponding in-domain test sets and the non-sarcastic subset of SarcasTraps. Precision, recall, and F1 are computed for the hate speech class. Results are reported as mean $\pm$ standard deviation over five random seeds.}
\label{tab_0}
\end{table*}

The results from Table~\ref{tab_0} indicate that the models do not undergo a general performance collapse when transferred to the non-sarcastic portion of the targeted external test set. We therefore use SarcasTraps (0) as the reference condition for examining the additional change associated with sarcasm. This comparison should not be interpreted as showing that the in-domain test sets and SarcasTraps (0) have identical difficulty or class distributions; its purpose is to establish that the models retain meaningful hate speech detection ability in the absence of sarcasm.

Table~\ref{tab_1} compares performance on the non-sarcastic reference condition, SarcasTraps (0), and the sarcastic subset, SarcasTraps (1). Unlike the preceding analysis, this comparison includes the zero-shot LLMs because both conditions are available for all evaluated models.

\begin{table*}[t!]
\centering
\scriptsize
\setlength{\tabcolsep}{3.2pt}
\renewcommand{\arraystretch}{1.15}
\resizebox{\textwidth}{!}{
\begin{tabular}{l c c c c c c c c}
\toprule
\multirow{2}{*}{\textbf{Model}}
& \multicolumn{4}{c}{\textbf{SarcasTraps (0), $n=278$}}
& \multicolumn{4}{c}{\textbf{SarcasTraps (1), $n=201$}} \\
\cmidrule(lr){2-5}
\cmidrule(lr){6-9}
& \textbf{Acc.} & \textbf{Prec.} & \textbf{Rec.} & \textbf{F1}
& \textbf{Acc.} & \textbf{Prec.} & \textbf{Rec.} & \textbf{F1} \\
\midrule

BERTurk$_{\text{SU-HS}}$
& 0.817{\scriptsize $\pm$ 0.009}
& 0.871{\scriptsize $\pm$ 0.016}
& 0.787{\scriptsize $\pm$ 0.011}
& 0.827{\scriptsize $\pm$ 0.007}
& 0.630{\scriptsize $\pm$ 0.026}
& 0.692{\scriptsize $\pm$ 0.034}
& 0.372{\scriptsize $\pm$ 0.063}
& 0.483{\scriptsize $\pm$ 0.060} \\

BERTurk$_{\text{Mayda}}$
& 0.812{\scriptsize $\pm$ 0.020}
& 0.838{\scriptsize $\pm$ 0.018}
& 0.818{\scriptsize $\pm$ 0.042}
& 0.828{\scriptsize $\pm$ 0.022}
& 0.597{\scriptsize $\pm$ 0.007}
& 0.604{\scriptsize $\pm$ 0.014}
& 0.404{\scriptsize $\pm$ 0.045}
& 0.483{\scriptsize $\pm$ 0.031} \\

XLM-RoBERTa$_{\text{SU-HS}}$
& 0.778{\scriptsize $\pm$ 0.028}
& 0.898{\scriptsize $\pm$ 0.020}
& 0.675{\scriptsize $\pm$ 0.046}
& 0.770{\scriptsize $\pm$ 0.034}
& 0.623{\scriptsize $\pm$ 0.013}
& 0.746{\scriptsize $\pm$ 0.024}
& 0.294{\scriptsize $\pm$ 0.030}
& 0.421{\scriptsize $\pm$ 0.032} \\

XLM-RoBERTa$_{\text{Mayda}}$
& 0.746{\scriptsize $\pm$ 0.027}
& 0.854{\scriptsize $\pm$ 0.034}
& 0.655{\scriptsize $\pm$ 0.041}
& 0.740{\scriptsize $\pm$ 0.031}
& 0.536{\scriptsize $\pm$ 0.020}
& 0.509{\scriptsize $\pm$ 0.035}
& 0.317{\scriptsize $\pm$ 0.009}
& 0.390{\scriptsize $\pm$ 0.014} \\

GPT-4.1 Zero Shot
& 0.878
& 0.826
& 0.987
& 0.899
& 0.622
& 0.566
& 0.819
& 0.670 \\

Llama-3.3-70B Zero Shot
& 0.802
& 0.744
& 0.981
& 0.846
& 0.567
& 0.526
& 0.755
& 0.620 \\

\bottomrule
\end{tabular}
}
\caption{Hate speech detection performance on the non-sarcastic and sarcastic subsets of SarcasTraps. Precision, recall, and F1 are computed for the hate speech class. Encoder results are reported as mean $\pm$ standard deviation over five random seeds. Zero-shot LLM results are from deterministic decoding with temperature 0.}
\label{tab_1}
\end{table*}

The degradation associated with sarcasm is systematic across all four encoder configurations. On SarcasTraps (0), the encoder models obtain mean F1 scores between 0.740 and 0.828. On SarcasTraps (1), their mean F1 scores fall to between 0.390 and 0.483. The absolute F1 decrease is highly consistent across the configurations, ranging from 0.344 to 0.350.

The encoder models' main limitation on sarcastic examples is low hate-class recall. Recall decreases from 0.655-0.818 on SarcasTraps (0) to 0.294-0.404 on SarcasTraps (1), an absolute reduction of 0.338-0.415. Moreover, recall remains below precision for every encoder configuration on the sarcastic subset. This indicates that the fine-tuned encoders frequently fail to retrieve hateful examples when group-directed hostility is expressed through a sarcastic surface form. Because the pattern occurs under both encoder architectures and both training datasets, it is not specific to a single configuration.

The zero-shot LLMs exhibit a different error profile. GPT-4.1's F1 decreases from 0.899 on SarcasTraps (0) to 0.670 on SarcasTraps (1), while Llama-3.3-70B's F1 decreases from 0.846 to 0.620. Both models retain comparatively high hate-class recall on sarcastic examples: 0.819 for GPT-4.1 and 0.755 for Llama-3.3-70B. However, their precision decreases from 0.826 to 0.566 and from 0.744 to 0.526, respectively. Their comparatively high recall therefore comes with substantially more false-positive hate predictions on non-hateful sarcastic examples.

To better understand the two main error patterns observed in the quantitative results, we examined two groups of misclassified instances in detail. For the encoder models, we examined hateful-sarcastic instances misclassified as non-hateful. Encoder errors were determined from the majority prediction across the five random seeds. For the zero-shot LLMs, we examined non-hateful-sarcastic instances misclassified as hateful. 

At least one encoder configuration missed 81 of the 94 hateful-sarcastic instances (86.2\%), while all four configurations missed 41 instances (43.6\%). 

Qualitative inspection revealed recurring errors involving indirect hostility expressed through apparent approval or mock praise, as well as implied threats or references to violence. These errors were not limited to seed-sensitive boundary cases.
Depending on the encoder configuration, 68.9-80.4\% of the hateful-sarcastic false negatives were misclassified by all five seeds. 
Moreover, 23 instances were assigned the non-hate label by every seed of all four encoder configurations. 

The LLMs showed the opposite tendency on non-hateful sarcasm. At least one LLM produced a false positive on 77 of the 107 non-hateful-sarcastic instances (72.0\%), and both LLMs misclassified 46 of these instances. Qualitative inspection suggested that the LLMs often misclassified non-hateful sarcastic political comments and criticism as hate speech. 

\section{Conclusion}
In this paper, we presented SarcasTraps: Effect of Sarcasm on Turkish Hate Speech Detection. In the set, each example is annotated for both hate speech and sarcasm, yielding four conditions that separate literal hate, non-hateful sarcasm, and sarcastically expressed hate. This design allows us to evaluate not only whether models detect hate speech in general, but also whether they remain reliable when harmful intent is expressed indirectly.

Experiments with BERTurk and XLM-RoBERTa classifiers trained on existing Turkish hate speech datasets, evaluated across five random seeds, as well as zero-shot LLMs, show that sarcasm exposes a systematic weakness in current hate speech detection systems. Fine-tuned encoder-based models tend to miss sarcastically expressed hate, while zero-shot LLMs retain higher recall on sarcastic hate but also produce more false positives on non-hateful sarcastic examples. These results indicate that sarcasm affects hate speech detection in different ways across model families and that standard evaluation settings can obscure these failures.

Overall, SarcasTraps demonstrates that sarcasm is not only a phenomenon to be detected on its own. It is also a factor that can affect model behavior in safety-critical tasks. It shows that robust hate speech detection requires models to distinguish not only harmful from non-harmful language, but also literal harmfulness from indirectly expressed group-directed hostility.

\section{Limitations}

SarcasTraps is designed as a targeted test set for evaluating the effect of sarcasm on Turkish hate speech detection systems. It contains 479 examples, which limits the scale of the quantitative evaluation. This size is partly due to the difficulty of finding naturally occurring examples that simultaneously involve hate speech and sarcasm. While hateful language and sarcastic language are both common in informal online discourse, examples where the two interact clearly enough to support reliable annotation are much rarer. For this reason, the results should be interpreted as evidence from a targeted test set.

Another limitation is the class distribution of the set. Because Ekşi Sözlük is organized around titles, data collection required identifying relevant titles and then selecting entries under those titles. To obtain enough hateful examples, we intentionally inspected titles where hate speech was more likely to occur, such as discussions involving socially or politically salient target groups. As a result, SarcasTraps contains more hateful than non-hateful examples. This imbalance reflects the targeted purpose of the dataset, but it also means that the set does not represent the natural prevalence of hate speech or sarcasm.

Future work can extend SarcasTraps by adding more examples and covering additional target domains beyond the ones considered in this study. A larger and more diverse benchmark would allow more fine-grained analyses. Another important direction is the development of sarcasm-aware hate speech detection systems. The results in this paper suggest that models trained on standard hate speech datasets are not sufficient for robustly handling sarcastically expressed hate, so future systems may benefit from sarcasm detection signals, or multi-task learning setups that jointly consider sarcasm and hate speech.

Finally, future work should examine whether context improves hate speech detection in sarcastic cases. Prior work showed that title-level context can help sarcasm detection in Ekşi Sözlük-style discourse. A natural next step is to investigate whether similar title-level or discussion-level context can also improve the detection of hateful intent when it is expressed indirectly through sarcasm. Such work could clarify whether failures on sarcastic hate are caused primarily by missing pragmatic cues in the entry itself, or by the absence of broader contextual information about the title, event, or target group being discussed.

Because SarcasTraps focuses on hate speech and informal online discourse, it contains sensitive and harmful language that is not censored in order to preserve the original content.

\section{Ethical Considerations}
This study aims to support research on sarcasm-aware hate speech detection in Turkish online discourse. Because SarcasTraps focuses on hate speech and informal online discourse, it contains sensitive and harmful language that is not censored in order to preserve the original content. 

SarcasTraps is constructed from naturally occurring entries on Ekşi Sözlük and is intended solely as a research resource for analyzing how sarcasm affects hate speech detection systems. The dataset should not be used to amplify hateful content or to target individuals or communities as the set itself may cause harm if redistributed without restrictions, or used outside of its intended research purpose.

All examples were collected by the authors. The examples were independently annotated by three native Turkish speakers: the two authors and an Turkish MSc student from the same laborator  y who is not an author. The annotators are native Turkish speakers who are familiar with Ekşi Sözlük discourse and the goals of the project. The annotators were not recruited through a crowdsourcing platform and did not receive payment.
We do not report protected demographic attributes of annotators, beyond task-relevant information such as native Turkish proficiency and familiarity with Ekşi Sözlük discourse.
Before collection and annotation, they were informed that the data would contain hateful, offensive, and disturbing content. During annotation, they focused only on two specific dimensions: whether an entry contains hate speech and whether it is sarcastic. They did not evaluate the political correctness, factual accuracy, moral acceptability, or ideological position of the entries.

Because the set was constructed from a public source, and the collection followed the platform's robots.txt restrictions, no formal ethics review was obtained for this study. We do not release the set as a public download; instead, access will be provided only upon request for academic-only purposes. 
The dataset contains only the title, entry content, hate label, and sarcasm label. We do not collect or release usernames, timestamps, profile information, or other metadata that could be used to identify Ekşi Sözlük authors to reduce privacy risks. We did not obtain individual consent from Ekşi Sözlük users, as the entries were collected from publicly accessible pages. We acknowledge that public data does not always translate to consent for research~\cite{fiesler2018participant}. As suggested in ~\citep{fiesler2018participant}, we are not mass gathering entries from specific people. The titles and entries in the set do not include any identifying information. We masked individual person names appearing in the entries, but all the other content remains uncensored. Due to sensitive nature of the data, SarcasTraps will be distributed only upon academic request under research-only terms.

Large language models were used as zero-shot classifiers in the experiments; they were not used to generate SarcasTraps entries or assign the annotations. We used ChatGPT 5.6 and Grammarly to check spelling and text flow after drafting our own version, and incorporated suggestions for more appropriate wording.

\section*{Acknowledgements}
This work was supported, in part, by Sabanci University, project number B.A.CF-25-03063. We would also like to thank Doruk Benli for his valuable contribution to the development of the SarcasTraps dataset.

\bibliography{references}

\appendix

\section{Zero-shot LLM Prompt for Hate Speech Classification}
\label{app:prompts}

We used the same Turkish prompt for zero-shot hate speech classification with GPT-4.1 and Llama-3.3-70B-Versatile. The models were instructed to return a binary JSON label and a short justification. The original Turkish system message is provided below.

\subsection{Turkish System Message}
\begin{Verbatim}[breaklines=true, breakanywhere=true, fontsize=\small]
Sen bir nefret söylemi (hate speech) sınıflandırıcısısın. 
Kullanıcı sana Türkçe bir metin verecek. 
Görevin: metnin nefret söylemi içerip içermediğine karar vermek.

Nefret söylemi; bir kişiyi veya grubu ırk, etnisite, din, cinsiyet, 
cinsel yönelim, engellilik veya benzeri kimlik özelliklerine dayanarak 
hedef alan, aşağılayan, tehdit eden veya şiddete teşvik eden içeriktir.

Yalnızca aşağıdaki JSON formatında yanıt ver, kod bloğu veya ek metin OLMADAN:
{ "label": 1 | 0, "reason": "<kısa gerekçe>" }

1 = nefret söylemi var, 0 = nefret söylemi yok.
Emin değilsen en olası etiketi seç ve bunu gerekçede belirt.
\end{Verbatim}

\subsection{English Translation of System Message}
\begin{Verbatim}[breaklines=true, breakanywhere=true, fontsize=\small]
You are a hate speech classifier.
The user will provide a Turkish text.
Your task is to decide whether the text contains hate speech.

Hate speech is content that targets, demeans, threatens, or encourages 
violence against a person or group on the basis of identity-related 
attributes such as race, ethnicity, religion, gender, sexual orientation, 
disability, or similar characteristics.

Respond only in the following JSON format, without code fences or extra text:
{ "label": 1 | 0, "reason": "<short reason>" }

1 = hate speech is present, 0 = hate speech is not present.
If uncertain, choose the most likely label and state this in the reason.
\end{Verbatim}

\section{User Prompt Template}
\label{app:user_prompt}

For each SarcasTraps instance, we formed the input text by concatenating the title and entry content as \texttt{title + " : " + content}, and inserted the resulting text into the following user prompt.

\begin{Verbatim}[breaklines=true, breakanywhere=true, fontsize=\small]
Aşağıdaki metni analiz et ve nefret söylemi içerip içermediğine karar ver.

Metin:
{{TEXT}}

Yalnızca şu JSON nesnesini döndür:
{ "label": 1 | 0, "reason": "<kısa gerekçe>" }
\end{Verbatim}

We provide the English translation below for clarity.

\begin{Verbatim}[breaklines=true, breakanywhere=true, fontsize=\small]
Analyze the following text and decide whether it contains hate speech.

Text:
{{TEXT}}

Return only the following JSON object:
{ "label": 1 | 0, "reason": "<short reason>" }
\end{Verbatim}

\section{LLM Output Parsing}
\label{app:llm_parsing}
For both models, we used JSON object response formatting and deterministic decoding with temperature 0. Outputs were parsed as JSON objects and the label was normalized to a binary value, where 1 indicates hate speech and 0 indicates no hate speech.

\end{document}